\documentclass[floats,superscriptaddress,reprint, pre]{revtex4-1}

\usepackage{here}
\usepackage{amssymb}
\usepackage{amsmath}
\usepackage{dsfont}
\usepackage{psfrag}
\usepackage[percent]{overpic}
\newcommand{\fsz}{\footnotesize}
\newcommand{\ssz}{\scriptsize}

\usepackage[T1]{fontenc}
\usepackage[latin9]{inputenc}
\usepackage{lmodern}
\usepackage{graphicx}
\usepackage[usenames,dvipsnames]{color}
\usepackage[english]{babel}
\usepackage{placeins}
\usepackage{comment}
\usepackage{booktabs}
\usepackage{fullpage}
\usepackage{url}
\usepackage{hyperref}
\usepackage{tikz}

\usepackage{lipsum}
\usepackage{babel}
\makeatother

\usepackage{color}

\begin{document}
\title{The Loss Surface of XOR Artificial Neural Networks}

\author{Dhagash Mehta}\email{mehtadb@utrc.utc.com}
\affiliation{Systems Department, United Technologies Research Center, East Hartford, CT, USA}

\author{Xiaojun Zhao}\email{zhaox@utrc.utc.com}
\affiliation{Systems Department, United Technologies Research Center, East Hartford, CT, USA}

\author{Edgar A.~Bernal}\email{bernalea@utrc.utc.com}
\affiliation{Systems Department, United Technologies Research Center, East Hartford, CT, USA}

\author{David J.~Wales}\email{dw34@cam.ac.uk}
\affiliation{University Chemical Laboratories, Lensfield Road, Cambridge CB2 1EW, UK}

\begin{abstract}
\noindent 
Training an artificial neural network involves an optimisation process over the
landscape defined by the cost (loss) as a function of the the network
parameters. We explore these landscapes using optimisation tools developed for
potential energy landscapes in molecular science. The number of local minima
and transition states (saddle points of index one), as well as the ratio of
transition states to minima, grow rapidly with the number of nodes in the
network. There is also a strong dependence on the regularisation parameter,
with the landscape becoming more convex (fewer minima) as the regularisation term increases.
We demonstrate that in our formulation, stationary points for networks with $N_h$ hidden nodes,
including the minimal network required to fit the XOR data, are also stationary
points for networks with $N_h+1$ hidden nodes when all the weights involving the additional
node are zero. Hence, smaller networks optimized to train the XOR data are
embedded in the landscapes of larger networks. Our results clarify certain
aspects of the classification and sensitivity (to perturbations in the input data)
of minima and saddle
points for this system, and may provide insight into dropout and network compression.
\end{abstract}

\maketitle

\section{Introduction}\label{Sec:Introduction}
In recent years, machine learning \cite{christopher2006pattern}, and
particularly deep learning \cite{lecun2015deep,bengio2015deep}, techniques have
proved to be highly effective in automating complex tasks. 
Applications include face and object recognition, scene
understanding, natural language processing, speech recognition, game playing,
stock-market analysis, and prognostic health management, among others.

A machine learning algorithm can be viewed as a functional
mapping between inputs and outputs, the parameters of the mapping being
tunable. The learning or training process involves
optimisation of the machine parameters
to minimize a cost (or loss) function. 
The loss surface describes the relationship between the
values of the parameters and a performance metric,
which may include
a regularisation term. The reference input-output data points used in the learning stage
comprise the training set. The performance metric measures the deviation between the network
output and the true output for a given input, and the regularisation term may be included
to reduce overfitting.

For some machine learning algorithms, such as linear regression, ridge
and lasso regression, \textit{etc.}, the loss surface is usually convex,
meaning that there is a single minimum, which simplifies
the optimisation task.
However, for more sophisticated machine learning techniques, most notably artificial
neural networks (ANNs) commonly encountered in deep learning, the
loss surface is non-convex \cite{bengio2015deep} with multiple minima. In fact, some studies have demonstrated that the
loss surface of single-neuron models may have an exponentially increasing number
of minima with increasing number of inputs (see, \textit{e.g.},~\cite{auer1996exponentially}).  
In \cite{blum1988training}, it is shown that
training a network with a single hidden layer with two hidden nodes, one output and $n$ inputs, is a non-deterministic polynomial-time
(NP)-complete problem.  
Commonly used iterative gradient descent-based
algorithms for optimisation can converge to local minima rather than the global minimum.
Different random initializations of the iterative optimisation process will lead to different local minima,
but are unlikely to locate the global minimum reliably unless the number of minima is small.

Despite the large number of local minima expected for highly non-linear and
non-convex functions, deep learning frameworks perform
reasonably well, as evaluated in terms of performance on previously unseen test data.
One of the possible explanations for
this observation is that there may be no `bad' minima at all
\cite{baldi1989neural,saxe2013exact,dauphin2014identifying,NguyenLoss,safran2016quality,goodfellow2014qualitatively,lin2016does,soudry2016no}, where 
`good' and `bad' minima are loosely defined in terms of the quality of the
network performance on the training set (\textit{i.e.}, the empirical error). In
\cite{coetzee1997488,dauphin2014identifying}, however, it was also argued that difficulties in reaching the global minimum in such models arise from
the proliferation of saddle points, and further \cite{dauphin2014identifying},
that saddle points give an illusion of a `bad' minimum because they correspond
to higher cost function values. This situation slows down learning, which is typically implemented using gradient-based methods, such as 
stochastic gradient descent \cite{bottou1998online,lecun2012efficient} rather than second-order methods. Alternative methods to escape
from saddle points have therefore been proposed, but many of them
eventually struggle to deal with the NP-hardness of the problem \cite{anandkumar2016efficient}. The large ratio of 
saddle points to minima is well known for molecular energy landscapes, 
both from theory \cite{2003JChPh.11912409W} and numerical investigations
\cite{2002JChPh.116.3777D,Mehta2011,Mehta:2011xs,Mehta:2013fza,mehta2015statistics,mehta2014potential}.

Recently, \cite{kawaguchi2016deep} showed that the loss surfaces
of deep \textit{linear} neural networks, (\textit{i.e.},~multilayer neural networks with
activation functions that are linear with respect to the input, but with
possibly non-linear loss functions) as well as those of certain special but
unrealistic cases of deep nonlinear networks, have degenerate global minima with 
the same value for the cost function.
Moreover, they also showed that for these networks, `bad' saddle points are indeed present, potentially
making the task of reaching a global minimum difficult.

In \cite{choromanska2014loss}, the loss surface of a fully-connected
feedforward deep network with one output and rectified linear (ReLU) activation
functions was approximated by the mean-field
spherical $p$-spin model of statistical physics
\cite{crisanti.1992,kurchan1993barriers,castellani2005spin}, to take advantage of
analytical results based on
random matrix theory \cite{fyodorov2014topology,fyodorov2013high,auffinger2013random,auffinger2013complexity,Mehta:2014xya,Mehta:2013fza}.
With this approximate model, the authors showed (and later further confirmed
\cite{sagun2014explorations}) that the number of saddle points with lower
indices, at which the value of the loss surface is beyond a certain threshold,
diminishes exponentially as the size of the deep network increases. The saddle points and minima at
which the value of the loss function is below the threshold were found to be
`good' minima. Hence, the authors concluded that the deeper the network, the
less likely it is to find `bad' minima. 

In \cite{BallardDMMSSW17} the ANN landscape was explored systematically using
optimisation tools developed in the context of molecular potential energy landscapes \cite{wales03} 
(see also \cite{pavlovskaia2015mapping,ballard2016energy} for other applications).
In particular, a three-layer ANN architecture was employed 
to fit the Modified National Institute of Standards and Technology (MNIST) data \cite{lecun1998mnist}, and the resulting landscapes for the cost function were found to
exhibit a single-funnel structure \cite{doyewm98,doyemw99a,wales03}.  This
organization is associated with efficient relaxation to the global minimum in molecular science,
and has been identified with `magic number' clusters,
crystallization, and protein-folding landscapes \cite{walesmw98,wales03,WalesB06}.
Although the number of local minima increases exponentially with the number of degrees of freedom,
relaxation to the global minimum is effectively guided downhill in energy (the
value of the loss function in an ANN landscape) over relatively low barriers \cite{wales03}.  
Ref.~\cite{BallardDMMSSW17} provides a review, highlighting the connections between
molecular potential energy surfaces and ANN landscapes.

In this paper, we present a detailed analysis of the landscape of the 
loss function for ANNs using tools employed in the analysis of 
energy landscapes. In particular, we aim to 
perform a (near) exhaustive search of minima, saddle points of index one (transition states \cite{murrelll68}), 
as well as pathways that connect pairs of minima via transition states. 
Another goal of this study is to investigate the recently debated quaestion of `bad' vs `good' minima in the deep 
learning literature, in the context of ANNs.

For our purposes the choice of data needs to satisfy certain
criteria: (1) the data should be non-trivial, \textit{i.e.}, they should not 
be linearly separable; (2) the architecture of the ANN to completely fit 
the chosen data should be known; (3) the ANN to completely fit the 
data should only have a small number of degrees of freedom, so that we can
obtain an essentially complete set of minima.
We considered various public datasets, including the well-known MNIST database
of hand-written digits.
However, these did not satisfy our criteria for the present investigation. Moreover, the current
best network for the MNIST dataset has far too many degrees of freedom (the number of network parameters) for  an
exaustive search of local minima to be practical. 
In contrast, the exclusive OR (XOR) data is both non-trivial (XOR is non-linearly separable) 
and simple enough to satisfy our criteria. Furthermore, the loss function of various ANNs that fit the XOR function 
has been studied in many previous contributions, as reviewed in the next section.
Hence, we focus on the XOR function in this paper.

Our principal results are: 
(1) a complete characterization of local minima (\textit{i.e.}, all the different possible trainings) as well as transition 
states and pathways between them; (2) a classification of
`good' and `bad' minima based on the sensitivity of the corresponding trained network to perturbation of the input data, as well as 
the sparsity (\S \ref{sec:loss}) of the trained network; (3) identification of `bad' minima, which exist even 
for the simple XOR function; (4) demonstration that as long as 
an ANN with one hidden layer is overspecified (\textit{i.e.}, the hidden layer has more neurons 
than the minimum number required to fit the data), the landscape contains 
an optimal ANN as a minimum of the loss function where all the incoming 
and outgoing weights for certain neurons vanish.

In the remaining sections we describe the XOR function, explain the set up of the problem, and provide a brief summary 
of the relevant results. 

\section{Loss Function of XOR}
\label{sec:loss}
The exclusive OR (XOR) between two boolean variables is a logical operation
whose output is true only when the two inputs have different values.  This
function has been the subject of several studies aimed at gaining insight into
properties of loss functions for small neural networks, because of the nonlinear separability of the data
\cite{Sprinkhuizen-Kuyper94theerror,Sprinkhuizen-Kuyper:1996:ESS:1362137.1362146,Hamey:1998:XNL:294223.294231,PMID:12662806,Sprinkhuizen-Kuyper1999,774274}.

The neural network in our study has one input layer with two nodes, one hidden
layer with $N_{h}$ nodes, and one output layer with two nodes.  The training
set comprises two inputs with four possible combinations $\textbf{X} = \{(0,0), (0,1), (1,0),(1,1)\}$ and their
corresponding outputs $\textbf{Y}=\{0, 1, 1, 0\}$. 
The two output nodes correspond to probabilities assigned by the network to possible output values 0 and 1.  For each
node $j$  in the hidden layer, a bias $\omega_{j}^{\rm bh}$ is added to the sum
of the corresponding weights used in the activation function. Similarly, for each node
$i$ in the output layer, a bias
$\omega^{\rm bo}_i$ is added to the sum of corresponding weights. Hence,
with the weights on the links between hidden node $j$ and output $i$ being
$\omega_{ij}^{(1)}$ and those on the links between input $k$ and hidden node
$j$ being $\omega_{jk}^{(2)}$, and with $\tanh$ as the activation function, the input-output relationship of the network in question is
\begin{equation}
y_i = \omega^{\rm bo}_i + \sum ^{N_{h}}_{j=1} \omega_{ij}^{(1)} \tanh\big(\omega^{\rm bh}_{j} + \sum_{k=1}^{2} \omega_{jk}^{(2)} x_k),
\end{equation}
where the internal sum from $1$ to $2$ corresponds to the two components of the input data.
The outputs are then converted into softmax probabilities using
\begin{equation}
p_c(\textbf{W}; \textbf{X}) = e^{y_c} / \left( e^{y_0}+e^{y_1}\right) .
\end{equation}

To train the model, we minimize 
\begin{equation}
\label{eq:loss}
E(\textbf{W}; \textbf{X}) = -\frac{1}{|\textbf{X}|} \sum_{d=1}^{|\textbf{X}|} \ln p_{c(d)} (\textbf{W}; \textbf{X}) + \lambda \textbf{V}^2, 
\end{equation}
with respect to  \textbf{W} = \{$\omega_{ij}^{(1)}, \omega_{jk}^{(2)},
\omega_{j}^{\rm bh}$, $\omega^{\rm bo}_i\}$, where
$c(d)$ is the correct outcome for data item $d$.
The loss function includes a performance and a regularisation term; 
$\textbf{V}$ is either a vector containing the $\omega^{(1)}_{i,j}$ and $\omega^{(1)}_{j,k}$ parameters, 
or alternatively all the parameters
$\textbf{V} = \textbf{W}$. We will return to the specific
choice of $\textbf{V}$ below. $|\textbf{X}| = 4$ is the number of
data points, which corresponds to the cardinality of the training input set,
and $\lambda$ is the regularisation parameter. In our experiments, we consider $\lambda = 10^{-l}$ for $l=1,2, \ldots ,6$. 
In Eq. (\ref{eq:loss}), cross-entropy is used as the performance term, where one of the probabilities in the true distribution
is one and the others are zero. Hence the summation reduces to a single term $-\ln p_{c(d)}$ for each data item.

Depending on the formulation, the XOR loss function may possess a range of discrete and continuous symmetries
\cite{sussmann1992uniqueness,chen1993geometry,badrinarayanan2015symmetry}. Hence, once a minimum or a
stationary point is found, others can be found having the same loss function value (infinitely many if continuous symmetries exist, corresponding to
zero eigenvalues of the Hessian second derivative matrix). 
In the present work these degenerate solutions are recognised by tight convergence of the loss function, and are lumped together.

For the loss function defined in Eq. (\ref{eq:loss}) it is straightforward to show that 
a stationary point for a network with $N_h$ hidden nodes is also a stationary point of some index for a network with $N_h+1$ hidden nodes
if all the weights involving the additional node are zero (see Appendix \ref{sec:appendix}). 
The extra degrees of freedom associated with the larger network introduce more flexibility,
which we would expect to lead to a lower value for the loss function after relaxation,
and perhaps to a higher Hessian index.
However, the regularisation term introduces a competing effect,
so the stationary point corresponding to the augmented network with zero weights for hidden
node $N_h+1$ could be a minimum or a saddle point.
Larger values of $\lambda$ will penalise non-zero values for the additional weights,
and are therefore more likely to conserve the Hessian index. We refer to the network in which at least one of the 
weights is zero as a sparse network; for a given network, the larger the number of zero weights, the sparser the network.

\subsection{Previous Analysis of the XOR Loss Landscape}
It was initially shown that one particular formulation of the loss function for the simplest network required
to solve the XOR problem has only one minimum
\cite{Sprinkhuizen-Kuyper94theerror,Sprinkhuizen-Kuyper:1996:ESS:1362137.1362146,Hamey:1998:XNL:294223.294231}. Ref. \cite{Sprinkhuizen-Kuyper1999} demonstrated the
absence of higher local minima in more complex networks (networks with two hidden
layers and two units in each layer) as long as the activation units are not
saturated \cite{PMID:12662806}; in contrast, when the activation units saturate
due to some weights having effectively infinite values, local minima start to appear. The existence
of suboptimal local minima in landscapes of more complicated networks (networks
with two hidden layers, two units in the first layer and three units in the
second layer) was demonstrated in \cite{774274}, directly contradicting earlier
assertions that two-layer neural networks with sigmoid activation functions and
$N_{h}-1$ hidden nodes do not have suboptimal local minima when the learning is
performed with $N_{h}$ training samples \cite{165604}. Ref.~\cite{coetzee1997488}
reported multiple minima for the XOR problem for specific neural network architectures. Recently, the authors of
\cite{swirszcz2017local} took a bottom-up approach in which they analyzed the
loss function of a simple (`flattened') neural network, which aims to approximate the XOR function, and found `bad' minima where learning fails. A thorough review of
the studies of 
optimisation landscape of the ANNs for XOR up to 2001 can be found in
\cite{gallagher2000multi} (see also \cite{rojas1996neural}).
We emphasise that the loss
functions in previous work are often constructed differently,
so the resulting landscapes may not be directly comparable.

Our inclusion of regularisation over all degrees of freedom in the present work is
a key difference from some previous studies, and simplifies the characterisation of
stationary points. Without sufficient regularisation, the machine learning landscape is
likely to include very flat regions, probably including non-Morse stationary points
with zero Hessian eigenvalues \cite{Hamey:1998:XNL:294223.294231}.
We explicitly wish to exclude such possibilities in the present analysis.

\section{Energy Landscape Theory and Computational Methods}
In chemical physics, the hypersurface defined by the potential energy
function, a multivariate nonlinear function of $3N$ atomic
coordinates for $N$ atoms, is referred to as the potential
energy landscape \cite{wales03}. The most interesting points
on the energy landscape are usually stationary points where
the gradient vanishes. Stationary
points are further classified according to the number of negative eigenvalues, or the index
$i$, of the Hessian (second derivative) matrix.
Stationary points of index $i=0$ are minima, where any small displacement of internal degrees of
freedom raises the energy.  
Local minima are connected by geometrically defined steepest-descent paths from transition states,
which are saddles of index one \cite{murrelll68}.
Non-Morse stationary points have zero eigenvalues that do not result from continuous
symmetries of the Hamiltonian.

The computational methods employed for geometry optimisation and construction
of connected networks of local minima and transition states are well
established, and a brief summary is provided here.  More details are available
in reviews \cite{wales03}, including a recent contribution that focuses on
machine learning landscapes \cite{BallardDMMSSW17}.

Global minima and a survey of low-lying local minima were obtained by basin-hopping global 
optimisation \cite{lis87,walesd97a,waless99}, using our {\tt GMIN} program \cite{GMIN}.
Here, steps are taken between local minima, obtained via random changes
to the coordinates of the current minimum in the chain (in our case, the network weights), with an acceptance criterion
based on the change in the loss function scaled by a fictitious temperature parameter.

To determine connections between local minima via transition states we first
run double-ended
searches between specific pairs using the doubly-nudged \cite{TrygubenkoW04}
elastic band \cite{HenkelmanJ00,HenkelmanUJ00} approach, which interpolates
between the end points via a sequence of images.  The images corresponding to
local maxima are then converged to transition states using the single-ended
hybrid eigenvector-following algorithm
\cite{munrow99,henkelmanj99,kumedamw01,ZengXH14}.  For each transition state,
the two connected minima are determined by calculating (approximately) the two
geometrically defined steepest-descent paths.  The OPTIM program \cite{OPTIM}
was used for all these calculations.

Although this methodology is well established, some additional effort was required
to tighten convergence criteria and ensure the accuracy of the pathways in terms of the
connectivity. These changes were necessary because the landscapes in
question support very soft degrees of freedom, even when regularisation is included over
all variables. For example, with a single hidden node, the minimum
with zero weights is connected via a transition state with weights that
are very small in magnitude, and the difference in the loss function is only $0.1425\times10^{-10}$.
All transition states were therefore checked using eigenvector-following with analytical
second derivatives, and the steepest-descent paths were obtained using a second-order 
method \cite{pagem88} after determining the displacement from the transition state that maximised the
decrease in loss function.
This procedure produced consistent connections for all the transition states obtained when 
they were located in alternative runs.

Additional checks were performed to ensure that stationary points with zero Hessian eigenvalues were
not included in the databases. Here, zero eigenvalues were first defined using a cutoff of $10^{-9}$,
which is about an order of magnitude less than the values observed for the smallest legitimate eigenvalues.
Changing the cutoff to $10^{-10}$ for $\lambda=10^{-6}$ and rerunning connection attempts between all pairs
of minima did not produce any additional stationary points.

When regularisation is applied to all variables, there are no zero Hessian eigenvalues
caused by continuous symmetries,
but singularities arise if the bias weights are not regularised.
These degrees of freedom could be treated by projection and eigenvalue shifting \cite{wales03}
using analytical expressions for the corresponding Hessian eigenvectors.
In molecular geometry optimisation, continuous symmetries arise from overall translation and
rotation \cite{wales03};
in neural networks, a uniform displacement in the $w_i^{\rm bo}$ has no effect on the probabilities.
The additional zero Hessian eigenvalues pose no problems for the 
custom LBFGS minimisation routine (a limited memory version of the quasi-Newton Broyden \cite{Broyden70}, Fletcher \cite{Fletcher70}, Goldfarb \cite{Goldfarb70},
Shanno \cite{Shanno70} BFGS procedure) employed in {\tt GMIN} \cite{GMIN}.
However, shifting and projection would be required to locate transition states and
construct disconnectivity graphs (see \S \ref{sec:DG}). 
We therefore restrict our landscape characterisations to loss functions
with regularisation applied to all the variables, where the connectivity is well defined.

The database of minima and transition states resulting from systematic connection
attempts between pairs of local minima constitutes a transition network \cite{RaoC04,NoeF08,pradag09,Wales10a}.
Various techniques have been described for refining such networks \cite{CarrTW05,StrodelWW07,WalesCKDSW09} (See \cite{JosephRCMW17,BallardDMMSSW17}.
In the present work we employed the PATHSAMPLE program \cite{PATHSAMPLE} to distribute OPTIM
jobs and organize the resulting output to expand the stationary point databases.  The overall approach is known as
discrete path sampling \cite{Wales02,Wales04}.

\subsection{Disconnectivity Graphs}
\label{sec:DG}
The energy landscape is a high-dimensional object, which usually cannot be
visualized effectively as a three-dimensional surface.
Instead, disconnectivity graphs \cite{BeckerK97, walesmw98, KrivovK02, EvansW03} provide a powerful approach for
understanding the organization, faithfully representing the barriers between local minima.
The vertical axis in a disconnectivity graph corresponds to the energy (cost or loss function),
and each branch terminates at the value for a local minimum.
At a regular series of threshold values, the local minima are grouped into disjoint superbasins \cite{BeckerK97},
whose members can interconvert via transition states without exceeding the threshold.
Branches join together at the threshold where they can interconvert.
The horizontal axis is usually chosen so that the branches are spaced out and do not overlap;
order parameters can also be employed to arrange the branches \cite{KomatsuzakiHMRJW05}, or to
color them.

\section{Results and Discussion}\label{sec:results}
In this section, we present our results and discuss their interpretation.
In particular, we consider the dependence between $N_h$ and
$\lambda$ and the nature of the resulting landscape, the relationship between
the complexity of the network and energy values, and introduce
an empirical analysis of network sensitivity to perturbation of the inputs. 

\subsection{Machine Learning Landscapes}
\label{sec:landscapes}
\begin{figure*}[h]
\psfrag{epsilon1}[cl][cl]{\ 0.05}
\psfrag{epsilon2}[cl][cl]{\ 0.00001}
\psfrag{0.00001}[cl][cl]{\ 0.00001}
\begin{tabular}{ccc}
\fbox{\begin{overpic}[width=0.3\textwidth,angle=0]{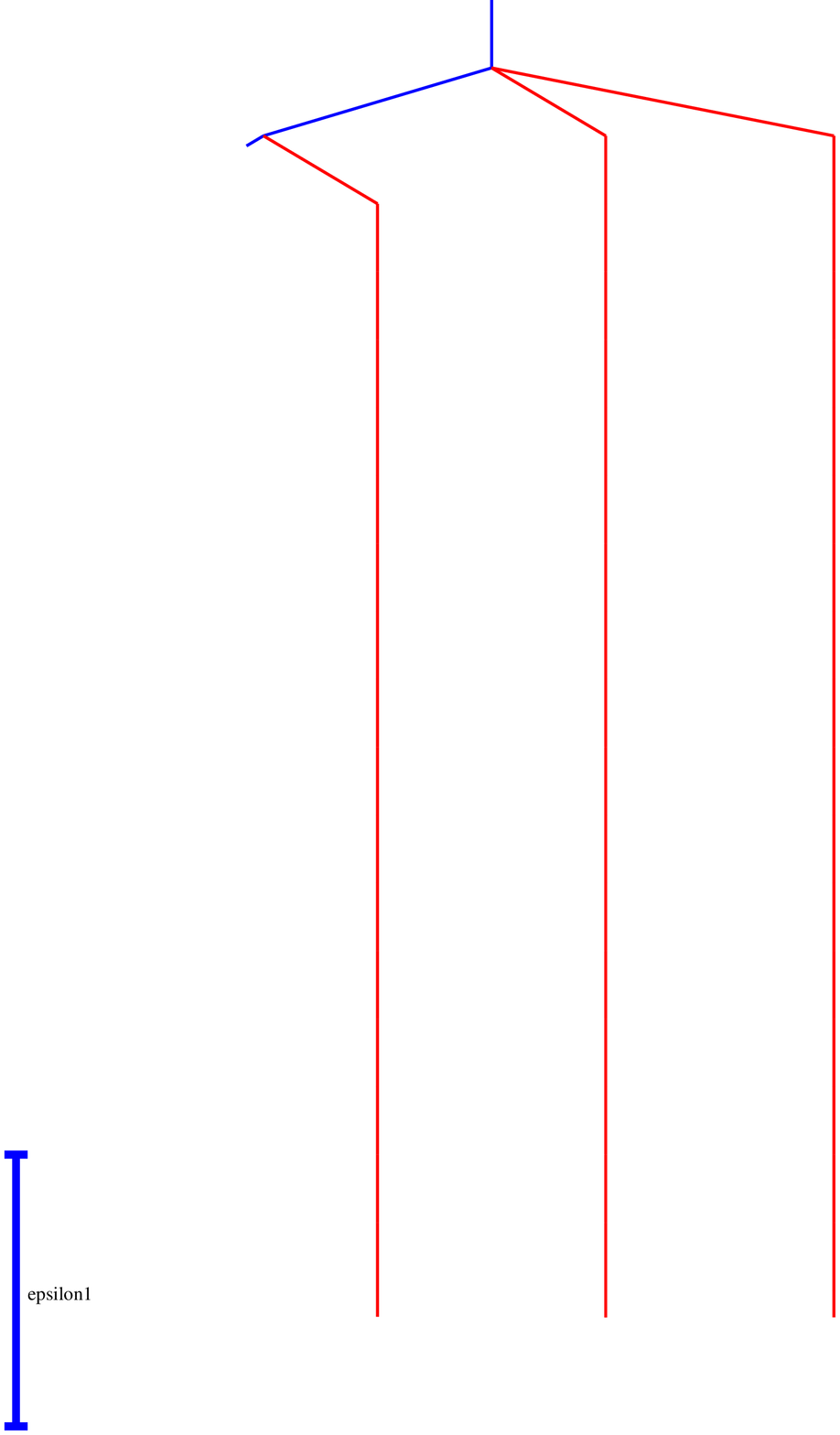}\put(5,85){(a)} \put(27,28){\ssz}\end{overpic}} &
\fbox{\begin{overpic}[width=0.3\textwidth,angle=0]{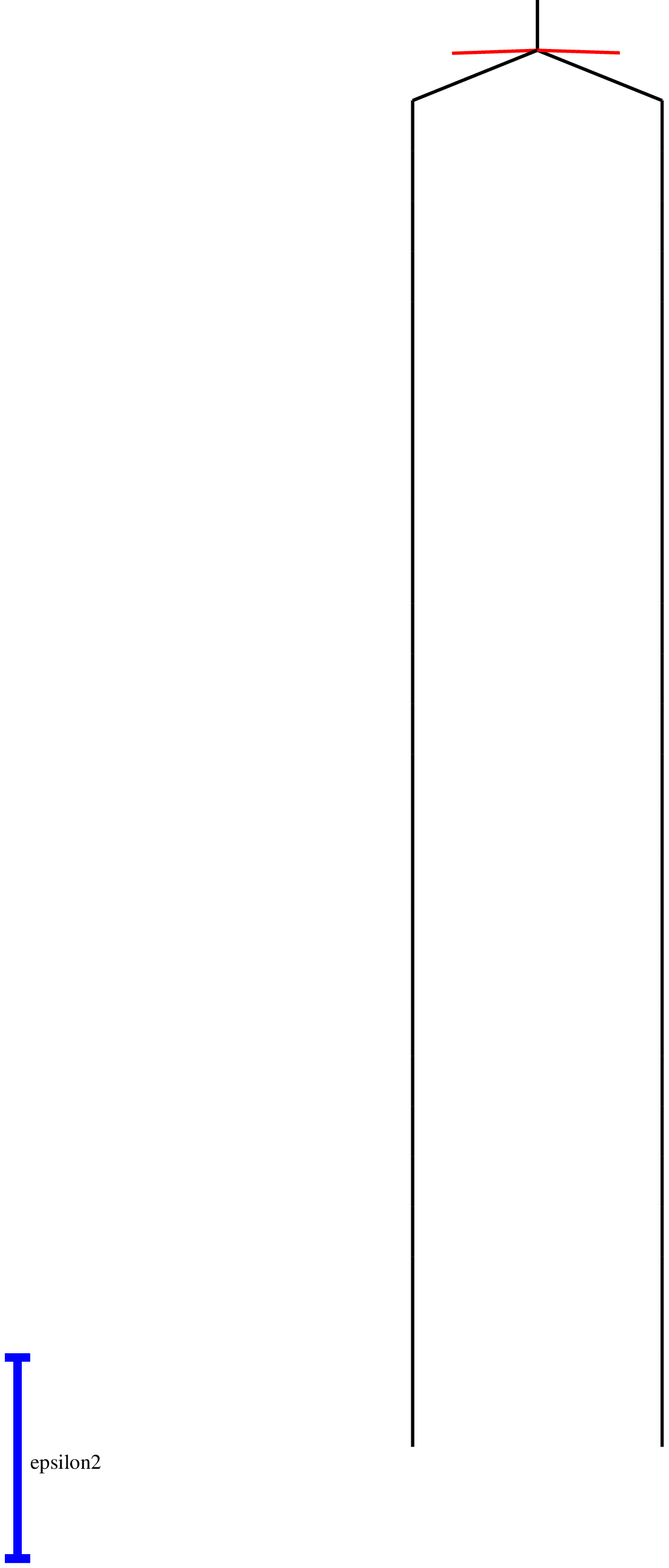}\put(5,85){(b)} \put(11,63){\fsz}\end{overpic}} &
\fbox{\begin{overpic}[width=0.3\textwidth,angle=0]{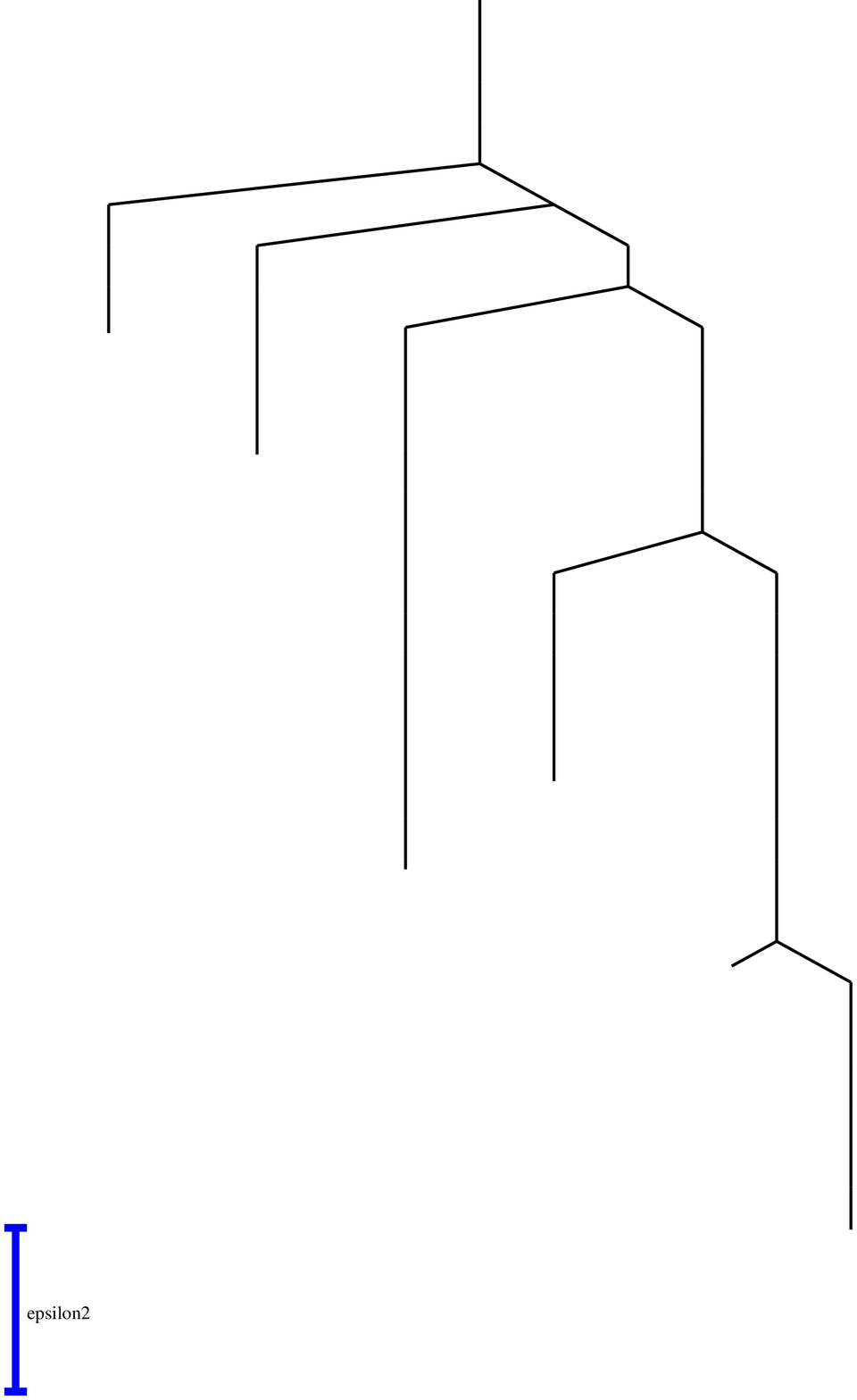}\put(5,85){(c)} \put(27,28){\ssz}\end{overpic}} \\
\fbox{\begin{overpic}[width=0.3\textwidth,angle=0]{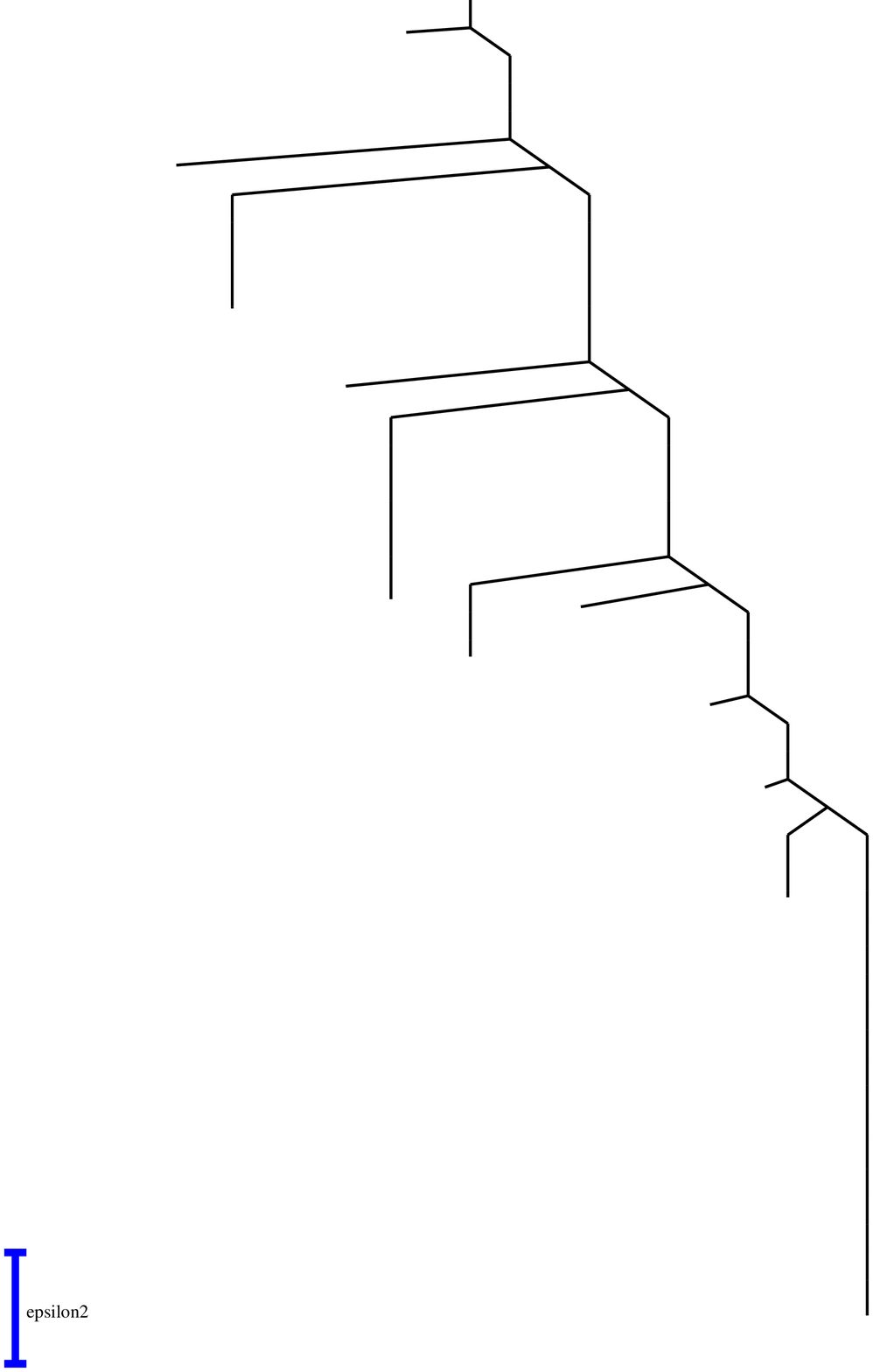}\put(5,85){(d)} \put(11,63){\fsz}\end{overpic}} &
\fbox{\begin{overpic}[width=0.3\textwidth,angle=0]{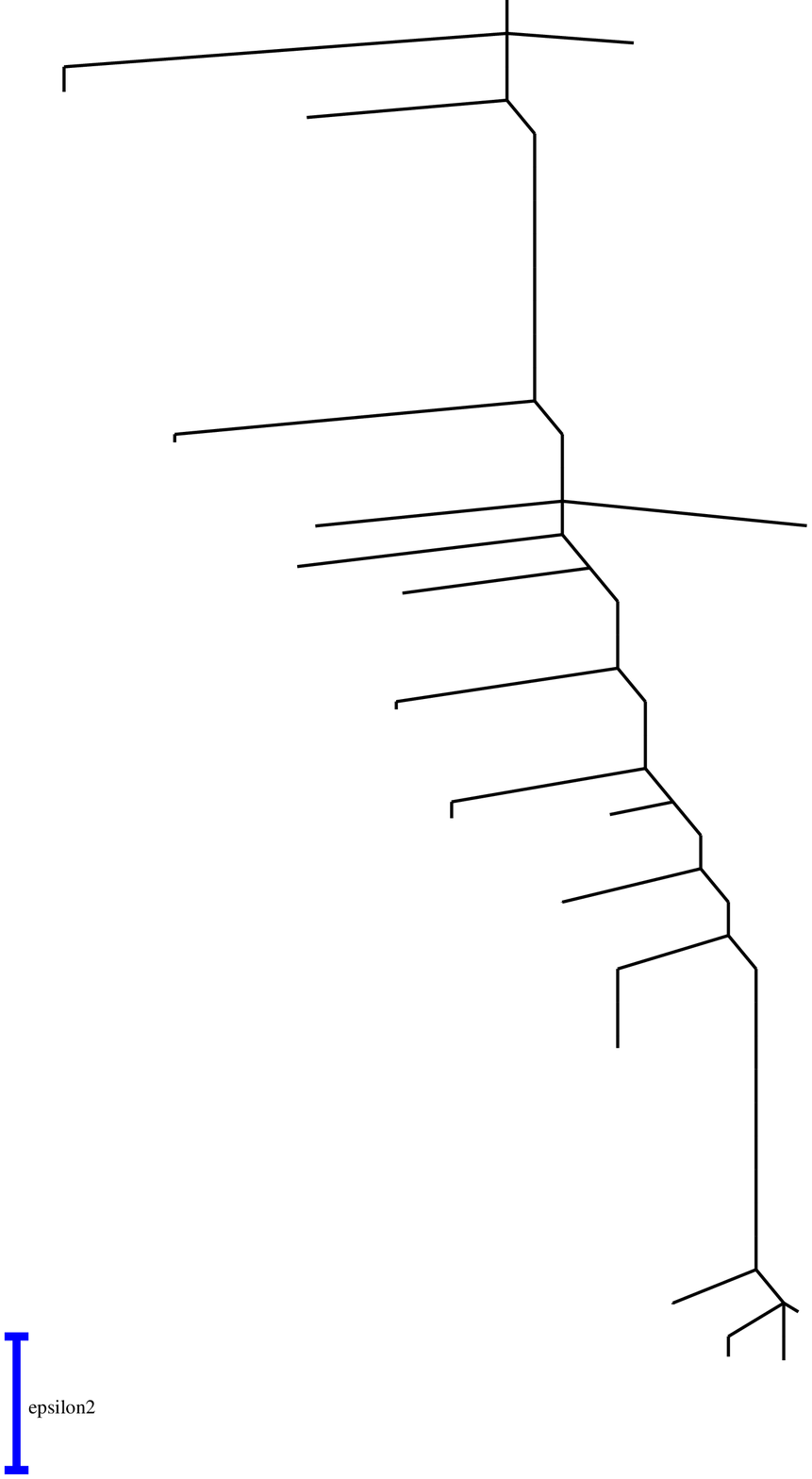}\put(5,85){(e)} \put(27,28){\ssz}\end{overpic}} &
\fbox{\begin{overpic}[width=0.3\textwidth,angle=0]{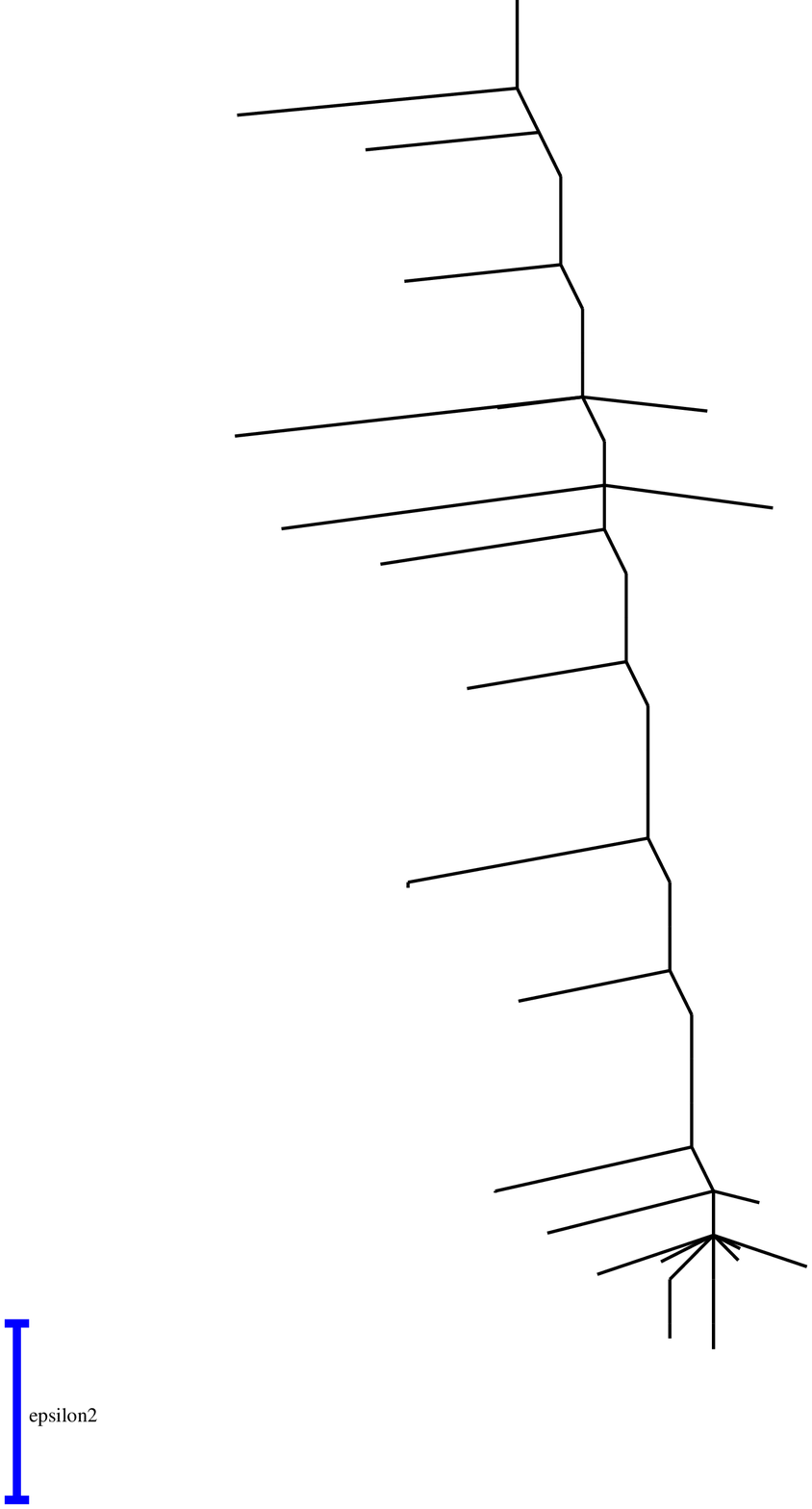}\put(5,85){(f)} \put(11,63){\fsz}\end{overpic}} \\
\end{tabular}
\caption{
Disconnectivity graphs for $\lambda = 10^{-6}$ obtained with neural networks fitted to the XOR function
containing $N_h=$ (a) $1$, (b) $2$, (c) $3$, (d) $4$, (e) $5$, and (f) $6$  hidden nodes. The blue bar 
illustrates the scale of the  vertical axis that represents the energy (loss function) values.
Note the contraction in scale by a factor of 5000 from 2 to 3 hidden nodes.
The branch coloured blue in panel (a) corresponds to a minimum with AUC 0.5; 
the branches coloured red in panels (a) and (b) correspond to minima with AUC values of 0.75.
All other minima have AUC values that are practically unity.
}\label{fig:lam0.000001}
\end{figure*}

\begin{figure*}[h]
\psfrag{0.01}[cl][cl]{\ 0.01}
\psfrag{0.001}[cl][cl]{\ 0.001}
\psfrag{0.0001}[cl][cl]{\ 0.0001}
\psfrag{0.00001}[cl][cl]{\ 0.00001}
\psfrag{0.0}[cc][cc]{0.0}
\fbox{\begin{overpic}[width=0.7\textwidth,height=0.85\textheight,angle=0]{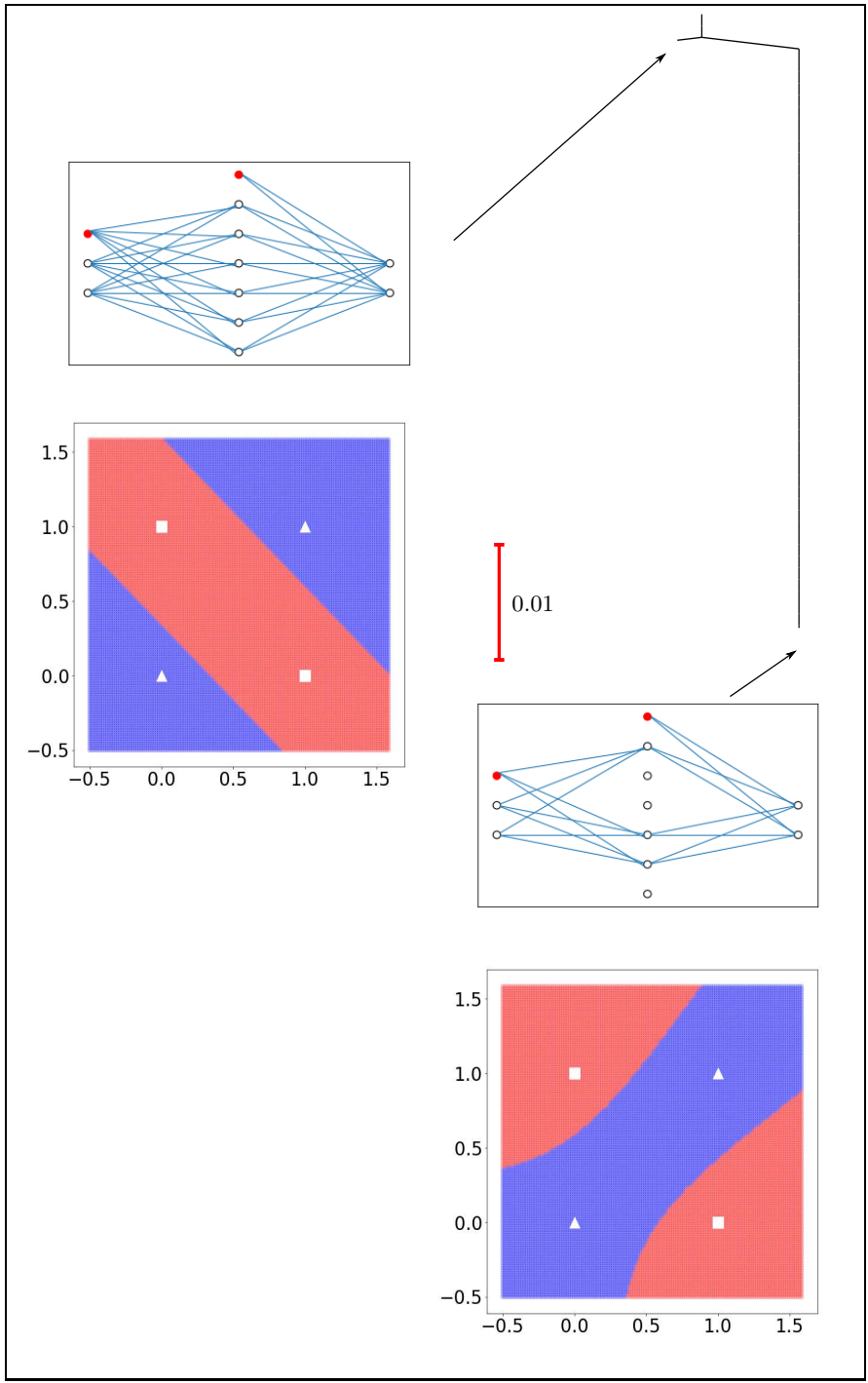}\put(11,63){\fsz}\end{overpic}}
\caption{Disconnectivity graph for the neural network containing 6 hidden nodes with regularisation parameter $\lambda = 10^{-2}$.
The figure also shows the neural network structures for the minima. The red colored nodes 
in the neural network diagrams are bias nodes. They are coloured red to distinguish them from the regular neurons.
An edge between two nodes corresponds to non-zero weights, whereas no edge means a numerically zero weight.
The figures below each network visualization illustrate the effect of perturbation to the inputs on the network output. 
The color of the data point with $(x,y)$ coordinates corresponds to the output
of the network for input values of $x$ and $y$: red and blue points correspond
to outputs $1$ and $0$, respectively. The white triangle and square symbols
represent the $0$ and $1$ output, respectively, for the actual inputs present
in the training set.}\label{fig:6_hidden_0pt01}
\end{figure*}

\begin{figure*}[h]
\psfrag{0.01}[cl][cl]{\ 0.01}
\fbox{\begin{overpic}[width=0.7\textwidth,height=0.85\textheight,angle=0]{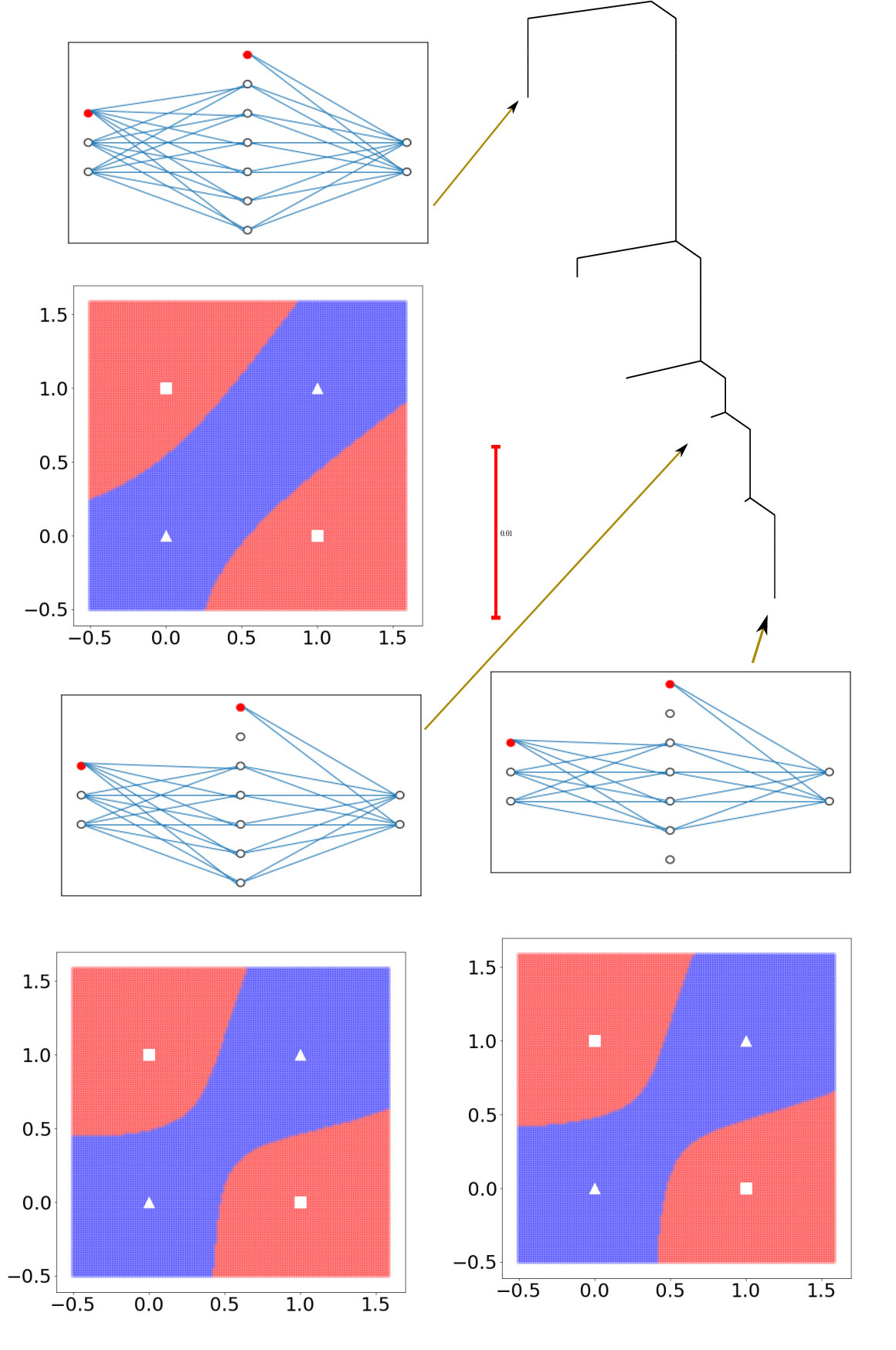}\put(11,63){\fsz}\end{overpic}}
\caption{Disconnectivity graph for $N_h = 6$ and $\lambda = 10^{-3}$. The layout of this figure is the same as 
for Fig.~\ref{fig:6_hidden_0pt01}.}\label{fig:6_hidden_0pt001}
\end{figure*}

\begin{figure*}[h]
\psfrag{0.001}[cl][cl]{\ 0.001}
\fbox{\begin{overpic}[width=0.7\textwidth,height=0.85\textheight,angle=0]{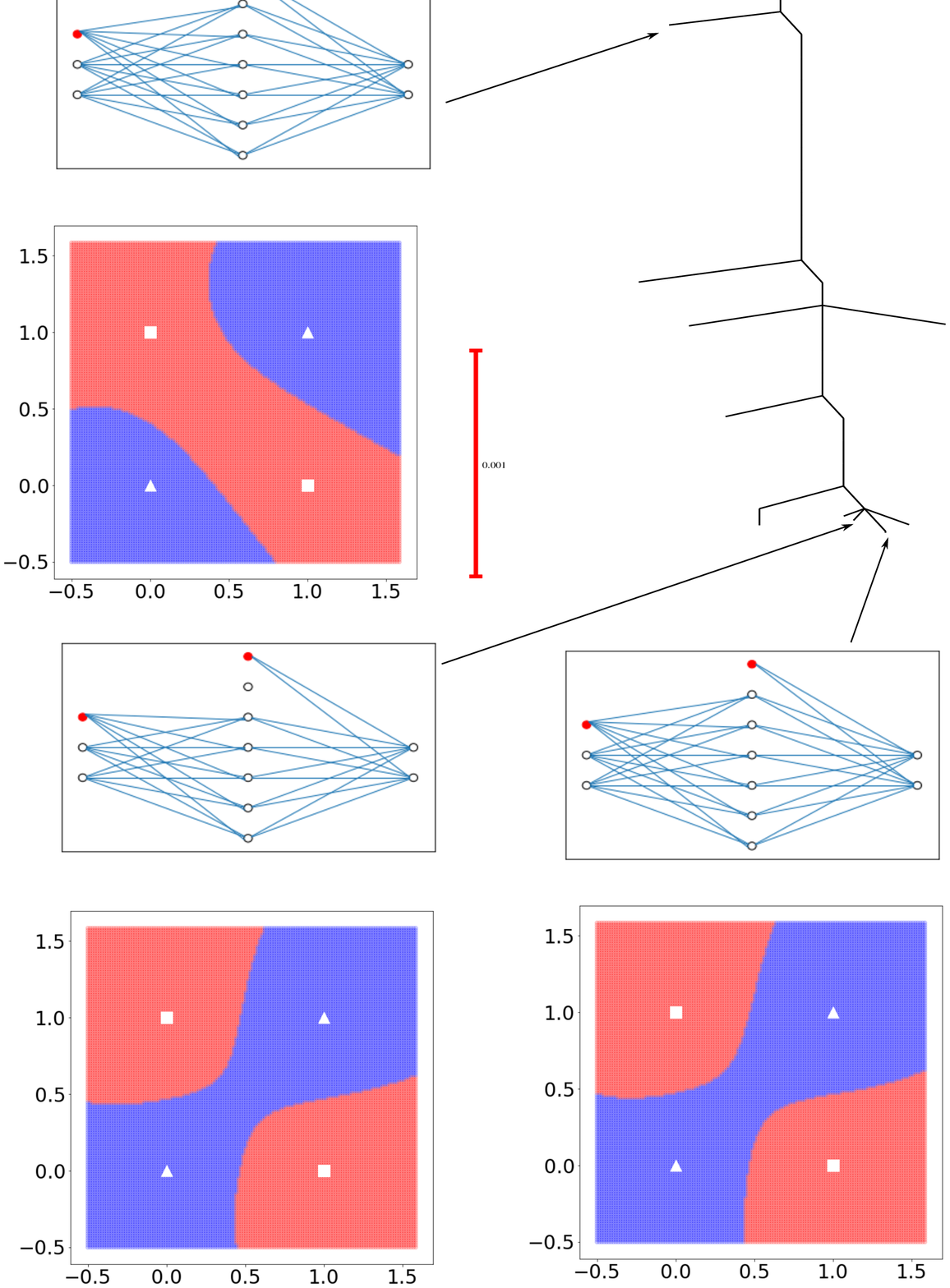}\put(11,63){\fsz}\end{overpic}}
\caption{Disconnectivity graph for $N_h = 6$ and $\lambda = 10^{-4}$. The layout of this figure is the same as 
for Fig.~\ref{fig:6_hidden_0pt01}.}\label{fig:6_hidden_0pt0001}
\end{figure*}

\begin{figure*}[h]
\psfrag{0.0001}[cl][cl]{\ 0.0001}
\fbox{\begin{overpic}[width=0.7\textwidth,height=0.85\textheight,angle=0]{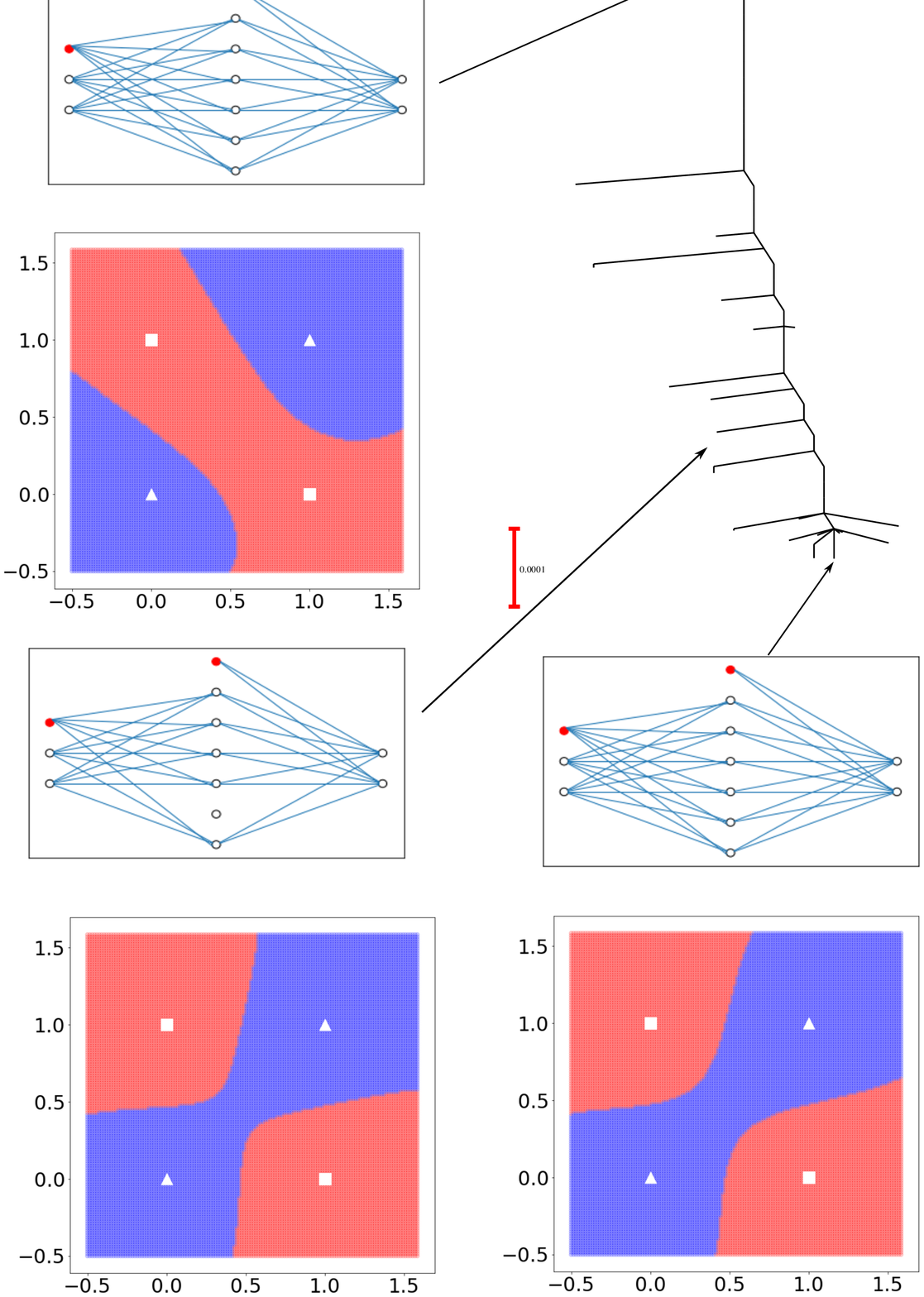}\put(11,63){\fsz}\end{overpic}}
\caption{Disconnectivity graph for $N_h = 6$ and $\lambda = 10^{-5}$. The layout of this figure is the same as 
for of Fig.~\ref{fig:6_hidden_0pt01}.}\label{fig:6_hidden_0pt00001}
\end{figure*}

\begin{figure*}[h]
\psfrag{0.00001}[cl][cl]{\ 0.00001}
\fbox{\begin{overpic}[width=0.7\textwidth,height=0.85\textheight,angle=0]{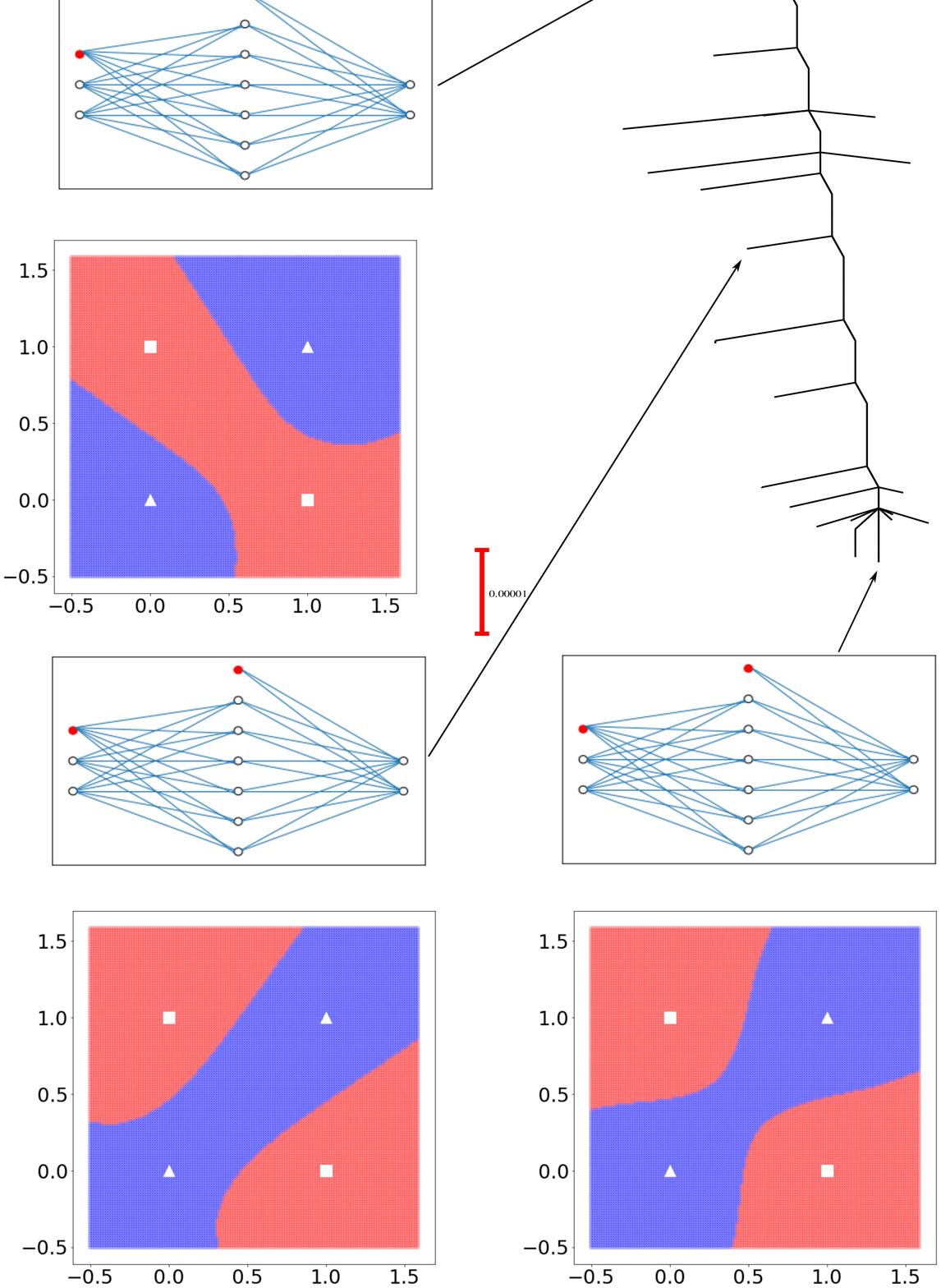}\put(11,63){\fsz}\end{overpic}}
\caption{Disconnectivity graph for $N_h = 6$ and $\lambda = 10^{-6}$. The layout of this figure is the same as 
for Fig.~\ref{fig:6_hidden_0pt01}.}\label{fig:6_hidden_0pt000001}
\end{figure*}

Machine learning landscapes were constructed for the loss function of Eq.~(\ref{eq:loss}) 
with various combinations of regularisation parameters and hidden nodes.
Selected examples are illustrated here using disconnectivity graphs \cite{BeckerK97, walesmw98, KrivovK02, EvansW03}.
Fig.~\ref{fig:lam0.000001} shows the landscapes obtained for $\lambda=10^{-6}$ with one to six hidden nodes,
and Figs \ref{fig:6_hidden_0pt01}--\ref{fig:6_hidden_0pt000001} show
how the landscape changes with $\lambda$ for six hidden nodes.
In these graphs, the terminus of each branch corresponds to the value of the minimised loss function,
with the global minimum at the bottom. Solutions with identical values are lumped together, so the 
branches that appear degenerate in Fig.~\ref{fig:lam0.000001} (a) and (b) actually correspond to
slightly different loss values.
For three hidden nodes and above, networks corresponding to all the local minima represented in Fig.~\ref{fig:lam0.000001} 
provide an accurate fit of the four input data points.
To quantify the prediction quality, we calculated the area under the curve (AUC) values for
receiver operating characteristic (ROC) plots.
All the AUC values obtained for the training set of four possible input values are close to
unity, except for the coloured branches in Fig.~\ref{fig:lam0.000001} (a) and (b).
These results show that once sufficient network parameters are included, increasing 
the number of hidden nodes results in additional solutions corresponding to
local minima with only slightly different loss function values. 

For any values of $N_h$ and $\lambda$, the trivial solution where all weights are
zero is always a bad minimum (as defined below). In some cases including this minimum in
the disconnectivity graphs changes the scale dramatically.
Hence, we omit this minimum in most of the graphs. Note that in \cite{haeffele2015global}, it 
was shown that if the network architecture consists of
parallel subnetworks where each subnetwork has a particular architecture defined by a specific
elemental mapping, a minimum at which all weights in one of the
subnetworks are zero is the global minimum. However,
this formulation is different from the present work.

Figs.~\ref{fig:6_hidden_0pt01}--\ref{fig:6_hidden_0pt000001} illustrate 
the effect of the regularisation parameter when applied
over all the variables, including the bias weights. $\lambda$ basically sets the vertical scale,
corresponding to the optimised loss function value, and determines how many minima the
machine learning landscape can support. All the local minima in these graphs correspond to
AUC values close to unity.
The predicted probabilities for the correct output corresponding to the two minima in 
Fig.~\ref{fig:6_hidden_0pt01} vary between 0.85 and 0.91. 
For minima in Figs.~\ref{fig:6_hidden_0pt001}--\ref{fig:6_hidden_0pt000001}
corresponding to $\lambda=10^{-3}$ and smaller, the probabilities are 0.99 or better in each case.

The variation in the number of minima and transition states as a function
of $\lambda$ for a fixed network architecture can be understood from
catastrophe theory. As $\lambda$ increases from zero, 
minima and transition states merge via fold catastrophes, and the remaining 
uphill barriers from lower to higher energy minima generally increase.
A detailed analysis for
surfaces parameterized by a single parameter (such as $\lambda$ in our case)
is provided in
\cite{wales2001microscopic,bogdan2004new}. 
The decreasing number of minima (\textit{i.e.}, increasing convexity) observed when varying $\lambda$ between
$10^{-6}$ and $10^{-2}$ is also known as
topology trivialization \cite{fyodorov2013high} in statistical physics, a
phenomenon that has been noted for various energy surfaces
\cite{fyodorov2014topology,kastner2011phase,mehta2012energy,mehta2013energy}
including ANNs \cite{chaudhari2015trivializing}.

\subsection{Visualizing Networks at Minima}
In the insets of Figs.~\ref{fig:6_hidden_0pt01}-\ref{fig:6_hidden_0pt000001}, we provide visualizations of 
the ANN at a few representative minima. We represent each minimum as a network:
if a weight \textit{at the minimum} is numerically zero (\textit{i.e.},
$10^{-10}$ or smaller), 
we do not include a connection between the corresponding nodes. The figures 
show that for $\lambda=10^{-2}$, for one of the two minima the network is fully
connected, whereas for the other minimum exactly three neurons are connected
and the other three are disconnected. For other values of $\lambda$, we find some minima at which zero, one, or two neurons are disconnected, 
whereas all minima for $\lambda=10^{-6}$ correspond to fully connected networks.

In the absence of separate training and testing data, the quality of the solutions corresponding to local minima in the learning process of a network has
usually been measured in terms of the network performance on the training
set at the weight values determined by the minima in question, \textit{i.e.},
the empirical error. 
Here, we quantify the quality of minima by considering
both the associated empirical error and the capacity of the resulting network,
or Vapnik-Chervonenkis (VC) dimension \cite{doi:10.1137/1116025}, which can be
intuitively interpreted as the number of tunable parameters in a neural
network.  According to the structural risk minimization principle, first
proposed by Vapnik \cite{DBLP:books/daglib/0097035}, the optimal minimum
corresponds to the model with the smallest combined empirical risk and
capacity.  While increasing model complexity is usually accompanied by
decreased empirical error, higher capacity can also lead to overfitting, or the
inability to generalize beyond the training set. We restrict our analysis 
to models with enough capacity to fit the training data, and note that models with less complexity than required will be
penalized by the performance term in the structural risk expression.

In our calculations, the performance metric is the AUC, and the complexity of the
model is coarsely measured as the number of non-zero weights in the network. In the XOR case, we find
two separate regions of $\lambda$ (at the discrete values we have chosen):
for $\lambda=10^{-1}$ and $\lambda=10^{-2}$, there exists at
least one `bad' minimum (in terms of empirical error), whereas for $\lambda =
10^{-3},\ldots, 10^{-6}$ no such solutions are found except for the trivial minimum
with zero weights.
However, for
some but not all of the `good' minima, many of the links have a number of 
weights that are practically zero, which indicates that less complex models suffice to successfully
classify the four data points in the training set. There is only one (up to the
discrete symmetries) minimum at $\lambda=10^{-2}$ in which only as many weights are non-vanishing as are needed to construct a minimal neural network
to fit the XOR function, \textit{i.e.}, a fully connected network with three hidden nodes. 
Hence, if we define a suboptimal minimum as one in which more than the necessary number of weights
have non-vanishing values (\textit{i.e.}, one at which more than the
necessary number of neurons remain connected), then there is only one optimal
minimum: the other minima are suboptimal. 

\subsection{Optimal Network Configuration}
The minimal network we obtained in our experiments, for any value of $N_h >2$,
has three fully connected hidden nodes (\textit{i.e.}, all incoming and outgoing
weights are non-zero for these three neurons). However, it is well-known that
the XOR data can be fitted with a network having $N_h=2$. Hence a configuration with only
non-zero weights that connect exactly two neurons to the inputs and outputs, in
addition to the bias weights, should be the `best' configuration
for networks with $N_h > 2$. In fact, this configuration is not a
minimum but is always a saddle point of index $N_h - 2$ 
for the values of $\lambda$ considered here. In other words, for
the XOR data with the present network set up, when the size of
the network is larger than the minimal network, the minimal network
configuration is a saddle point. This conclusion
differs from a recent study \cite{sankar2017saddles} which showed that deep
networks converge to saddle points at
which the Hessian matrix is singular, because in our case the `best' saddle point is
not degenerate. Details of this computation will be discussed elsewhere.

\subsection{Network Sensitivity to Perturbations in the Input Data}
Neural networks are susceptible to so-called adversarial
examples, where small perturbations in the input can cause misclassification
\cite{DBLP:journals/corr/GoodfellowSS14}. For networks operating on
high-dimensional spaces, effective adversarial perturbations need to be
carefully engineered. In the present work
the effect of perturbations can be analysed in detail given the
low dimensionality of the input space. The sensitivity of a trained network to perturbations 
in the input data is also 
referred to as
\textit{stability} of the network \cite{kawaguchi2017generalization} (see, \textit{e.g.}, \cite{li2017visualizing}, for 
a recent attempt to relate generalization and geometry of landscapes of ANNs).
To this end, the output of the
network to inputs in the range $[-0.5, 1.5]$ with a step size of $0.015$ was
computed. The insets of
Figs.~\ref{fig:6_hidden_0pt01}-\ref{fig:6_hidden_0pt000001} located below the
network visualization of each of the selected minima illustrate the results.
The color of the data point with $(x,y)$ coordinates corresponds to the output
of the network for input values of $x$ and $y$: red and blue points correspond
to $1$ and $0$ outputs, respectively. The white triangle and square symbols
represent the $0$ and $1$ output, respectively, for the actual inputs present
in the training set. 
Intuitively, a stable network should output $0$ when the inputs are similar to
each other and $1$ otherwise, regardless of the actual values.
Specifically, and given the choice of the binary coding scheme $(x,y) =
\{(0,0), (0,1), (1,0), (1,1)\}$, for inputs satisfying $|x-y| \leq 0.5$, the
output of the network should be $0$, with a desired output of $1$ for every
other input combination.  Inspection of
Figs.~\ref{fig:6_hidden_0pt01}-\ref{fig:6_hidden_0pt000001} reveals that 
the sparser networks are more robust, although the converse is not always true.

\section{Conclusion and Outlook}
Using the energy landscape theory developed in chemical physics, we 
have investigated the optimisation landscape of the loss function for neural networks trained to approximate the XOR function.
Our network has one hidden layer with $N_h$ neurons in the hidden layer. 
We find that the number of minima and saddle points of index 1 change rapidly
with $N_h$ and regularisation parameter $\lambda$.
More importantly, we discovered that the loss surface includes minima 
where some of the weights are essentially zero (around $\sim 10^{-10}$ or below) such that some of the 
hidden neurons effectively appear disconnected. The number of disconnected neurons 
can vary with $\lambda$. In particular, for the XOR data and for any $N_h > 3$, there is always a minimum at which exactly three hidden neurons remain connected, with
all the remaining $N_h - 3$ neurons being disconnected, indicating that a network with $3$ hidden neurons is a minimal configuration to successfully separate 
the XOR data, whereas a $2$-neuron configuration (which can also separate the data) is found to be a saddle point of index $N_h - 2$ rather than a minimum.

The universal approximation theorem states that a feed-forward ANN with one hidden layer and finite $N_h$ 
can approximate any continuous function defined on compact subsets of the real space. The theorem assumes mild criteria on the activation function. 
All these criteria are satisfied by the ANN and the activation function (hyperbolic tangent) 
we have chosen in the present work. The theorem, however, does not yield a procedure to 
obtain the optimal number $N_h$ to approximate the function. At the very least, 
determining the optimal number would require a priori knowledge of
the data, for example, how much data is available, 
whether the data includes all the representative cases of the complete dataset, 
the amount of noise, \textit{etc.} Our study does not address the question of how to find the optimal number of hidden neurons analytically.
However, we have shown empirically for the XOR example that if we select a network 
with a larger number of neurons than the optimal number required to fit the data, then for a certain value of the regularization parameter 
there will be a minimum in the landscape at which the corresponding
network will have only the optimal number of neurons connected and all the others effectively disconnected.

Our results may also explain why a large number of network parameters are
found redundant in previous studies
\cite{denil2013predicting,denton2014exploiting}: if a model with more
parameters than needed to fit the data is used, a minimum with sufficient
regularisation will make the unwanted network parameters redundant during
minimization. Our approach in turn provides a systematic way
to compress networks, a topic that has recently attracted significant attention
for fast and low-power mobile applications \cite{han2015deep,kim2015compression} (see also ref.~\cite{lawrence1998size}
for an earlier attempt to use an optimization approach to reduce the number of network parameters, and \cite{reed1993pruning} for a review 
on other methods to prune ANNs).

We note that linear regression with $L_2$-regularisation guarantees a unique
minimum at which the unimportant features are removed from the model,
whereas the landscape of ANNs with $L_2$-regularisation consists of
multiple minima with differing numbers of zero weights.
Our results
indicate that scanning the landscape may provide a systematic way to
perform partial hyperparameter parameterization of ANNs, meaning
optimisation of the number of hidden layers and the number
of neurons in each layer. {Verifying this result for more complex datasets as well as for deeper ANNs may shed further light on the the 
more complex nature of the learning process. An effort to investigate this issue is in progress.}

Recent empirical findings on regularisation techniques aimed at decreasing
coadaptation across different neurons (\textit{e.g.}, dropout
\cite{DBLP:journals/corr/abs-1207-0580} and spatial dropout
\cite{DBLP:journals/corr/TompsonGJLB14}) indicate that one of the reasons
complex models have a tendency to overfit the training data is that,
on training, some units may learn to correct mistakes made by other
units. This compensation may be effective at improving performance on the training set, but
usually leads to poor generalization capabilities.  Our results reinforce this
hypothesis, since simpler models showcase smaller risk for coadaptation, as
there are fewer potential symbiotic relationships among neurons available
at training.

Finally, we empirically demonstrated that sparse networks tend to exhibit improved
(\textit{i.e.},reduced) sensitivity to perturbations in the input, as
evidenced by the $0$-valued responses around the diagonal defined by
$x=y$. In contrast, for other minima, undesired regions of $1$-valued responses occur around the $y=1-x$ diagonal.
This trend was observed across a range of values for the regularisation
parameter $\lambda$. This analysis quantitatively distinguishes between good
and bad training: 
if the learning is not done carefully, then it could lead to a network corresponding to a minimum in the loss function that may 
appear good enough for the specific data, but could be sensitive to perturbations in the inputs. Such a system may be more vulnerable to 
adversarial attacks. Whether such minima survive when more hidden layers are added is an open issue and should be further investigated.

In the future, we plan to devise an algorithm that directly finds the best
minimum in the minimal network sense (and the corresponding value of the
regularisation parameter), and extend these investigations to deeper
networks and larger datasets, which may help to resolve potentially more complex landscapes and issues 
concerning zero eigenvalues of the Hessian for the cost function, which result in `flat' minima \cite{keskar2016large,dinh2017sharp}.

\section*{Acknowledgement}
DM, XZ and EAB acknowledge internal funding from UTRC. DJW acknowledges
financial support from the EPSRC. We thank  Jose-Miguel Pasini, Kishore
Reddy, Kunal Srivastava and Amit Surana for their feedback.

\bibliographystyle{unsrt}

\newpage

\appendix
\section{Proof of Results in Sec.~\ref{sec:loss}}\label{sec:appendix}

Here, we prove that for the loss function defined in Eq. (\ref{eq:loss}),
a stationary point for $N_h$ hidden nodes is also a stationary point of some index for $N_h+1$ hidden nodes
if all the weights involving the additional node are zero. 

Let ${\bf W}^{N_h}$ be a stationary point for the network with ${N_h}$ hidden nodes, 
and ${\bf W}^{N_h+1}$ be the vector of weights for $N_h+1$ hidden nodes containing the
same weights as ${\bf W}^{N_h}$ augmented by zero entries for the additional node. 
For this choice we see that
\begin{eqnarray}
&& y_i({\bf W}^{N_h+1};{\bf X})  = \nonumber \\
&& \omega^{\rm bo}_i + \sum ^{N_h+1}_{j=1} \omega_{ij}^{(1)} \tanh\big(\omega^{\rm bh}_{j} + \sum_{k=1}^{2} \omega_{jk}^{(2)} x_k) \nonumber \\
&=& y_i({\bf W}^{{N_h}};{\bf X}) + \omega_{i N_h+1}^{(1)} \tanh\big(\omega^{\rm bh}_{ N_h+1} + \sum_{k=1}^{2} \omega_{ N_h+1k}^{(2)} x_k) \nonumber \\
&=& y_i({\bf W}^{{N_h}};{\bf X}). \\
&& {\rm Hence} \qquad  p_c({\bf W}^{N_h+1};{\bf X})=p_c({\bf W}^{{N_h}};{\bf X}) \nonumber \\
&& {\rm and} \qquad E({\bf W}^{N_h+1};{\bf X})=E({\bf W}^{{N_h}};{\bf X}) \qquad {\rm as\ well,} \nonumber 
\end{eqnarray}
since $\lambda |{\bf W}^{N_h+1}|^2=\lambda |{\bf W}^{N_h}|^2$ under these conditions.
Similar results follow for the first derivatives, where
\begin{align}
& \frac{\partial E(\textbf{W}; \textbf{X})}{\partial {\bf W}} = -\frac{1}{|\textbf{X}|} \sum_{d=1}^{|\textbf{X}|} \frac{1}{p_{c(d)} (\textbf{W}; \textbf{X}) }
\frac{\partial p_{c(d)} (\textbf{W}; \textbf{X})}{\partial {\bf W}} \nonumber \\
& + 2\lambda \textbf{W}, \\
& {\rm with} \qquad \frac{\partial p_{c(d)} (\textbf{W}; \textbf{X})}{\partial {\bf W}} 
= p_{c(d)} (\textbf{W}; \textbf{X}) \times \nonumber \\
& \left(
     \frac{\partial y_{c(d)} (\textbf{W}; \textbf{X})}{\partial {\bf W}} - \sum_{k=1}^{2} p_{k} (\textbf{W}; \textbf{X}) \frac{\partial y_{k} (\textbf{W}; \textbf{X})}{\partial {\bf W}} \right). \nonumber
\end{align}
For ${\bf W}^{N_h}$ a stationary point of the network with ${N_h}$ hidden nodes we have
\begin{align}
& \frac{\partial y_i (\textbf{W}^{N_h+1}; \textbf{X})}{\partial w^{(2)}_{jk}} = w^{(1)}_{ij} x_k {\rm sech}^2\left( w^{\rm bh}_j + \sum_{k=1}^{2} w^{(2)}_{jk} x_k \right) \nonumber \\
& = \begin{cases}
\frac{\displaystyle \partial y_i (\textbf{W}^{{N_h}}; \textbf{X})}{\displaystyle \partial w^{(2)}_{jk}} =0 & 1\le j \le {N_h}, \nonumber \\
      \displaystyle 0, & j=N_h+1.
\end{cases} \nonumber \\
& \frac{\partial y_i (\textbf{W}^{N_h+1}; \textbf{X})}{\partial w^{(1)}_{mj}} = \delta_{im} {\rm tanh}\left( w^{\rm bh}_j + \sum_{k=1}^{2} w^{(2)}_{jk} x_k \right) \nonumber \\
& = \begin{cases}
\frac{\displaystyle \partial y_i (\textbf{W}^{{N_h}}; \textbf{X})}{\displaystyle \partial w^{(1)}_{mj}} =0 & 1\le j \le {N_h}, \nonumber \\
      \displaystyle 0, & j=N_h+1.
\end{cases} \nonumber \\
\end{align}
Similary for the bias weights 
\begin{align}
& \frac{\partial y_i (\textbf{W}^{N_h+1}; \textbf{X})}{\partial w^{\rm bh}_{j}} = w^{(1)}_{ij} {\rm sech}^2\left( w^{\rm bh}_j + \sum_{k=1}^{2} w^{(2)}_{jk} x_k \right) \nonumber \\
& = \begin{cases}
\frac{\displaystyle \partial y_i (\textbf{W}^{{N_h}}; \textbf{X})}{\displaystyle \partial w^{\rm bh}_{j}} =0 & 1\le j \le {N_h}, \nonumber \\
      \displaystyle 0, & j=N_h+1.
\end{cases} \nonumber \\
& \frac{\partial y_i (\textbf{W}^{N_h+1}; \textbf{X})}{\partial w^{\rm bo}_{m}} = \delta_{im}  
= \frac{\displaystyle \partial y_i (\textbf{W}^{{N_h}}; \textbf{X})}{\displaystyle \partial w^{{\rm bo}}_{m}} =0 .
\end{align}
It is straightforward to show that the first derivatives also vanish for the regularisation term
at the corresponding stationary points.

\newpage
\section*{Figures}
\vfill

\newpage

\vfill

\newpage

\vfill

\newpage

\vfill

\newpage

\vfill

\newpage

\vfill

\end{document}